\documentclass[conference]{IEEEtran}
\IEEEoverridecommandlockouts
\usepackage{cite}
\usepackage{amsmath,amssymb,amsfonts}
\usepackage{algorithmic}
\usepackage{graphicx}
\usepackage{textcomp}
\usepackage{xcolor}

\usepackage{url}

\usepackage{tikz}
\usetikzlibrary{positioning}
\usepackage{algorithm}
\usepackage{subfig}

\def\BibTeX{{\rm B\kern-.05em{\sc i\kern-.025em b}\kern-.08em
    T\kern-.1667em\lower.7ex\hbox{E}\kern-.125emX}}
\begin{document}

\title{\textbf{\LARGE Electric Network Frequency Optical Sensing Devices}}
\author{\IEEEauthorblockN{Christos Moysiadis, Georgios Karantaidis, Constantine Kotropoulos}
\IEEEauthorblockA{Department of Informatics, Aristotle University of Thessaloniki
			\\ Thessaloniki, 54124, Greece \\
Email: \{mousiadi, gkarantai, costas\}@csd.auth.gr}
}
%

\maketitle

\begin{abstract}
Electric Network Frequency (ENF) acts as a fingerprint in multimedia forensics applications. In indoor environments, ENF variations affect the intensity of light sources connected to power mains. Accordingly, the light intensity variations captured by sensing devices can be exploited to estimate the ENF. A first optical sensing device based on a photodiode is developed for capturing ENF variations in indoor lighting environments. In addition, a device that captures the ENF directly from power mains is implemented. This device serves as a ground truth ENF collector.
Video recordings captured by a camera are also employed to estimate the ENF. The camera serves as a second optical sensor. The factors affecting the ENF estimation are thoroughly studied. The maximum correlation coefficient between the ENF estimated by the two optical sensors and that estimated directly from power mains is used to measure the estimation accuracy. The paper's major contribution is in the disclosure of extensive experimental evidence on ENF estimation in scenes ranging from static ones capturing a white wall to non-static ones, including human activity.
\end{abstract}


\section{Introduction}
\label{sec:intro}
The widespread and ever-increasing use of social media and the vast amount of video recordings shared online have exposed individuals to perpetrators seeking illicit profits. A plethora of tools can alter and manipulate digital content, as well as metadata information. Through editing, the content and the time the recording was captured can be altered for fraudulent purposes.

The Electric Network Frequency (ENF) signal is employed as an authentication signature in a wide range of multimedia applications, starting from audio \cite{Grigoras2005} and proceeding to video, e.g., \cite{Hajj2016,Vatansever2017,Nagothu2019,Karantaidis2021b}.
The ENF is embedded in digital content captured by microphones, devices plugged into power mains, or near power sources.
It fluctuates around its fundamental frequency at $50$ Hz in Europe and $60$ Hz in the U.S. due to the instantaneous differences between the consumed and the produced electric load in the power grid.
The ENF can also be found in the higher harmonics of the fundamental frequency \cite{Grigoras2007}.
A lot of effort has been paid to develop ENF estimation algorithms \cite{Nicolalde2009, Huijbregtse2009,Ojowu2012,Bykhovsky2013,Hajj2013,Hua2016,Skodras2016,Hajj2019,Karantaidis2021a}, tested mainly for audio forensics applications.

The application domain of ENF-related authentication research has been shifted to images and video.
It has been found that ENF can be embedded in video recordings through the intensity variations of different light sources.   A systematic study of ENF estimation in digital video recordings captured by various optical sensors was presented in \cite{Garg2013}.
An algorithm for detecting the presence of  ENF  was proposed in \cite{Vatansever2017}.
Simple Linear Iterative Clustering (SLIC) was employed to derive superpixels. It was attested that the algorithm could operate in short clips regardless of the camera sensor type. Extension of \cite{Vatansever2017} was presented in \cite{Karantaidis2021b}, allowing to handle both static and non-static video recordings.
An analytical model for video recordings captured by a rolling shutter mechanism was proposed in \cite{Vatansever2019}. A novel method of ENF estimation in video employing a rolling shutter mechanism was developed in \cite{Ferrara2021}. There, parametric and non-parametric spectral estimation approaches were combined to estimate accurately the ENF.
The rolling shutter mechanism was modeled in \cite{Su2014a}, resorting to multirate signal processing theory.

In this paper, two optical devices are developed to measure the ENF signal in indoor environments illuminated by two different light sources. The intensity emanating from light sources fluctuates due to the ENF variations in the power grid.
Taking advantage of this fact, the first optical sensor based on a  photodiode measures the light intensity fluctuations. To evaluate the accuracy of the ENF signal captured by the photodiode-based sensor, a ground truth ENF is essential. For this reason, a device that records the ENF signal directly from power mains is also implemented.
Luminance variations were also recorded in an indoor environment employing a common CMOS-based GoPro Hero8 camera, which was set to record at 30 fps at a resolution of 1080p  with the anti-flicker effect turned off to resemble the operation of a surveillance camera. The camera serves as a second optical sensor.
Comparing the ENF extracted from the photodiode against that estimated by the GoPro camera,  useful conclusions are drawn regarding their effectiveness.
The experiments involved various setups, such as varying video duration and distance between the optical sensor and the white wall background and introducing moving objects or human activity. The maximum correlation coefficient (MCC) between the ENF estimated by either optical sensor and that estimated from power mains was employed as a metric to assess the ENF estimation accuracy.

The paper's major contribution is in the disclosure of extensive experimental evidence, which demonstrates that under certain conditions, the photodiode sensing device delivers a reliable reference (i.e., ground truth) ENF signal, extending the work in \cite{Garg2013}. This could benefit practitioners and find use in real-life applications where it is quite difficult to acquire ground truth ENF through a Frequency Disturbance Recorder (FDR).
Several spectral analysis methods were employed for ENF estimation, such as the Short-Term Fourier Transform (STFT), the Blackman-Tukey (BT) spectral estimate, the Estimation of Signal Parameters with Rotational Invariant Techniques (ESPRIT) \cite{b30}, as well as the spectrum combining approach \cite{Hajj2013}.

The remainder of the paper is as follows. Section \ref{sec:Fundamentals} briefly describes ENF fundamentals. Section \ref{sec:Development} details the design and implementation of sensing devices. Section \ref{sec:res} discusses  ENF estimation. Section \ref{sec:conc} concludes the paper and suggests future research topics.

\section{ENF Fundamentals}
\label{sec:Fundamentals}

The ENF is embedded in video recordings captured by optical sensors in indoor environments illuminated by fluorescent lamps, halogen lamps, or incandescent bulbs.
In particular, light intensity fluctuates at twice the fundamental frequency of ENF, i.e., $100$ Hz in Europe and $120$ Hz in the U.S.

The low sampling rate of optical sensors results in severe aliasing and, consequently, hinders ENF estimation. To determine the aliased frequencies and estimate the ENF signal in video recordings, one should apply the sampling theorem \cite{Oppenheim2009}. The aliased frequency $f_{A}$ emanated from halogen/incandescent illumination is given as   $ f_{A} = |\hat{f}_{N} - \gamma \: f_{s}| \leq \dfrac{f_{s}}{2}$,
where $f_{s}$ denotes the sampling frequency of the camera, $\hat{f}_{N}$ is the frequency of light source illumination, and $\gamma$ stands for an integer factor \cite{Hajj2015}.

The STFT is employed for ENF estimation following the procedure in \cite{Karantaidis2021b}. That is, for  a window $w(t)$  $L$ samples long centered around $lG$, where $G$ is the hop size in samples, the Discrete-Time Fourier transform of the $l$th segment of $x(t)$ is computed \cite{Allen1977}:
\begin{equation}
X_{l}(\omega) = \sum^{\infty}_{t=-\infty} {x}(t) \: w(t-lG) \: \exp\left( -j\omega t \right).
\end{equation}

The BT spectral estimate is a refined periodogram  \cite{b30}:
\begin{equation}\label{BT}
\hat{\phi}_{BT}(\omega)=\sum^{M-1}_{\zeta=-(M-1)} w(\zeta)\: \hat{r}(\zeta) \: e^{-j \omega \,\zeta}
\end{equation}
where $\hat{r}(\zeta) = \dfrac{1}{N} \sum_{t=\zeta+1}^{N} x(t)x(t-\zeta)$, $-(M-1) \leq \zeta \leq M-1$, is the standard biased estimate of ${r}(\zeta)$  and ${w}(\zeta)$ is an lag-window of even symmetry.

The spectrum combining approach delivers an accurate power spectrum based on a weighted summation of multiple spectral bands from around the signal harmonics\cite{Hajj2013}. Each band's local signal-to-noise ratio (SNR) is employed to calculate the corresponding weights. The weighted summation is given by:
\begin{equation}
S(\omega)=\sum_{z=1}^{Z_a} w_{z} \: \phi_{BT}(z \, \omega)
\end{equation}
where $\phi_{BT}(z \, \omega)$ is the $z$th harmonic scaled power spectrum,  $w_{z}$ weighs the harmonic spectral bands around each harmonic taking into consideration the local SNR, and $Z_a=7$ denotes the number of harmonics considered.

ESPRIT was also used to estimate the ENF signal. The size of the sample covariance matrix was set to 10 $\times$ 10, and the line spectrum model was set to 3. Due to lack of space, the interested reader is referred to~\cite{Karantaidis2021b} \cite{b30}.

\section{Devices Design and Implementation}
\label{sec:Development}
To collect light intensity fluctuations using a photodiode, following the paradigm in \cite{Garg2013}, a first circuit was designed based on the photodiode BPW21. This photodiode was chosen due to its excellent light and spectral sensitivity in the visible range. A common and robust way to amplify the photocurrent generated by a photodiode is to utilize two operational amplifiers with a positive and a negative feedback loop to stabilize its behavior and produce a linear correlation between the light intensity and the current generated \cite{Hernandez2008}. The operational amplifiers OP07D were used as current-to-voltage converters to capture the fluctuations and pass them to the laptop sound card. The voltage from the power mains was dropped to 18 Volts AC using a center-tapped transformer. Also, a full bridge rectifier and a voltage regulator pair (LM7805/7905) were used to convert the AC voltage to DC to power the operational amplifiers. The next step was to wire appropriately the two amplification stages of the operational amplifiers. A combination of positive and negative feedback occurred in the first stage to amplify and clean the initial signal collected from the photodiode. In contrast, the signal was further amplified in the second stage by incorporating another negative feedback stage.

A second circuit was implemented to extract the ENF from the power mains. A transformer lowered the power mains AC voltage from 220 Volts to 18 Volts. A voltage divider further decreased the voltage to about 3 Volts peak-to-peak in accordance with the laptop sound card voltage tolerance \cite{Triantafyllopoulos2016}. The next stage included a high pass filter to eliminate the DC component with a cutoff frequency of 32 Hz. Finally, an anti-aliasing filter stage was deployed to control the cutoff frequency set at the Nyquist frequency. More specifically, since we wanted to sample the signal at 1 KHz, we set the value of $R4$ to 33 K$\Omega$ to filter the signal at around 500 Hz. The ENF extracted from power mains served as reference ENF. In practice, the reference ENF is recorded by an FDR, which provides accurate measurements up to $\pm \,5\cdot 10^{-4}$ Hz \cite{Ojowu2012}.

\begin{figure}[!ht]
  \centerline{\includegraphics[width=86mm,scale=0.8]{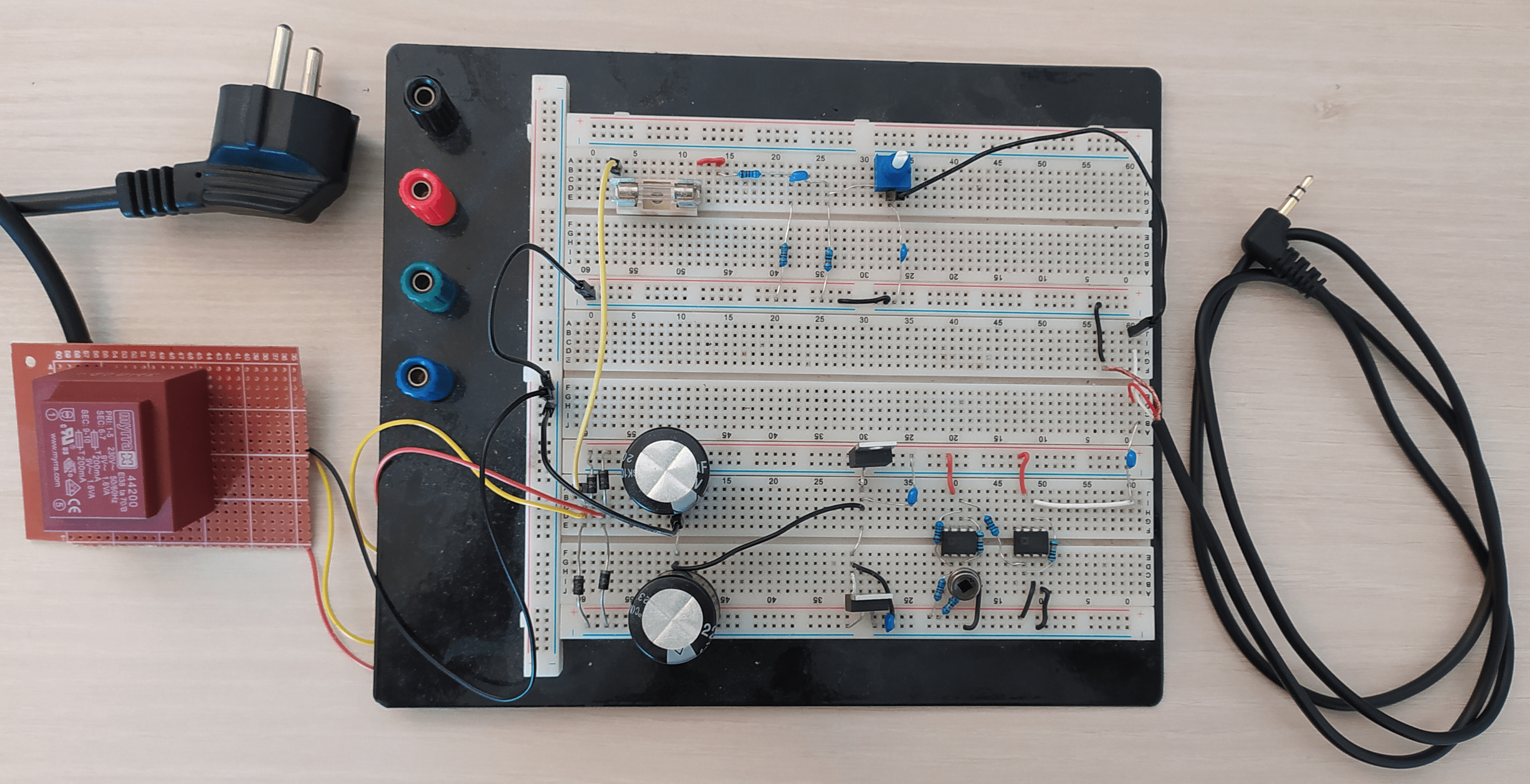}}
  \caption{\small Circuit photograph.}
  \label{circuit}
\end{figure}
To record simultaneously the ENF mains signal and the light intensity signal at a proper voltage level, we employed a 3.5 mm stereo jack cable to feed the ENF mains signal to the left channel of the sound card and the light intensity signal from the photodiode to the right channel of the sound card.
The developed devices are depicted in Figure~\ref{circuit}. On the left, the transformer of the circuit is depicted. The breadboard containing the circuit to extract the ground truth ENF from the power mains is shown on the top. The photodiode circuit, the full bridge rectifier, and the voltage regulation stages can be seen at the bottom. On the right, the 3.5 mm audio jack is depicted.

\begin{figure}[!ht]
  \centerline{\includegraphics[width=\linewidth]{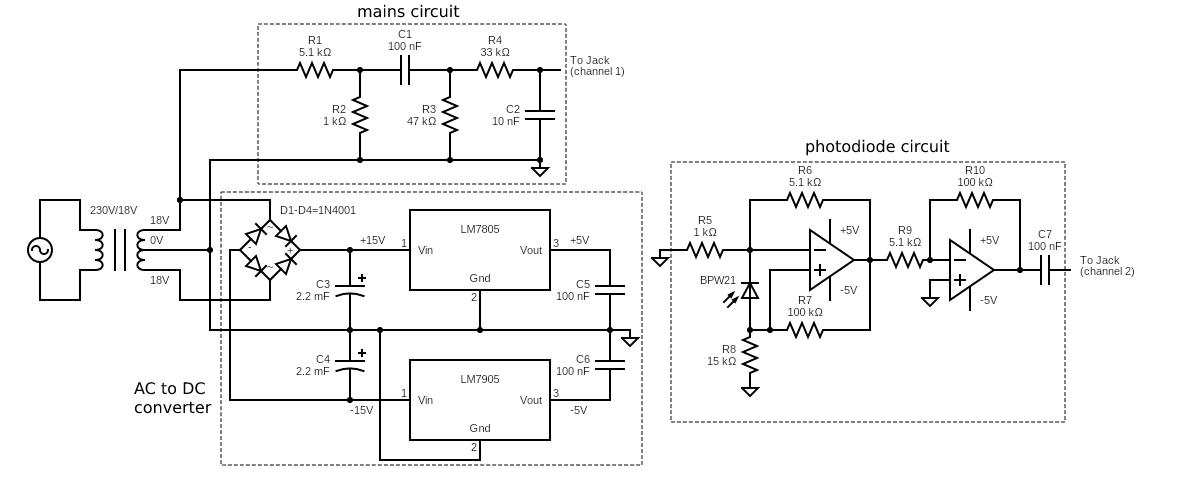}}
  \caption{\small Detailed circuit diagram of the developed devices.}
  \label{enf_circuit_diagram_annotated}
\end{figure}
The diagram of the devices is shown in Figure~\ref{enf_circuit_diagram_annotated}. On the left, a 230/18V center-tapped transformer is shown. The top comprises a voltage divider and two filtering stages for acquiring the power mains voltage. The central part of the diagram depicts a full bridge rectifier followed by a voltage regulator pair to convert the AC signal to +5/-5V DC rails, which are fed into the operational amplifiers. The photodiode and its two-stage amplification circuit can be seen on the right.
A MATLAB script \cite{enf_rec} was written to capture both signals.

\section{Device Experimental Evaluation}
\label{sec:res}

Two sets of experiments were conducted to evaluate the performance of the designed devices.

In the first set of experiments, a GoPro Hero 8 camera was employed to record a white wall inside a living room. Three different factors were taken into consideration: (i) two different light sources, namely, a halogen lamp and an incandescent bulb; (ii) different distances between the camera and the wall; and (iii) various video recording durations. In the second set of experiments, the same camera was used to collect video recordings under realistic conditions. Video recordings, ground truth ENF, and code can be found at~\cite{Video_GT_Code}.

\subsection{First set of experiments}
To assess the impact of the light source on the accuracy of ENF estimation, camera recordings of a white wall illuminated by either a halogen lamp or an incandescent bulb were collected.
Unless otherwise stated, the ENF estimation from video recordings was carried out using the SLIC-based approach~\cite{Vatansever2017}~\cite{Karantaidis2021b} and employing the STFT. Comparisons were made against the ENF estimated from the power mains using STFT.
The camera was mounted at different distances from the white wall background, namely 0.5 m, 1 m, 1.5 m, 2 m, 2.5 m, 3 m, and 3.5 m. The varying distances are found to affect the ENF  estimation, as can be seen in Table~\ref{dist}.
When a halogen lamp illuminated the scene, the top measured MCC was 0.9778 at a 1 m distance between the camera and the white wall. When an incandescent bulb illuminated the scene, the top measured MCC was $0.9738$ at the same distance.

\begin{table}[!htb]
\centering
\caption{\sc {\small  MCC Between the ENF Estimated from Video and Power Mains for Varying Distances}}
\label{dist}
\resizebox{0.55\columnwidth}{!}{%
\begin{tabular}{l|ll}
\hline \hline
\textbf{Distance} (m) & \textbf{Halogen} & \textbf{Incandescent} \\  \hline
0.5   &  0.8271  & 0.9232     \\
1   & \textbf{0.9778}   & \textbf{0.9738} \\
1.5 & 0.9058            & 0.8235 \\
2   & 0.9746            & 0.9394 \\
2.5 & 0.9210            & 0.9512\\
3   & 0.8266            & 0.7510 \\
3.5 & 0.7646            & 0.8525\\
 \hline \hline
\end{tabular}
}
\end{table}

\begin{figure}[!t]
   \centering
  \begin{tikzpicture}
  \node (img1)  {\includegraphics[width=85mm,scale=0.9]{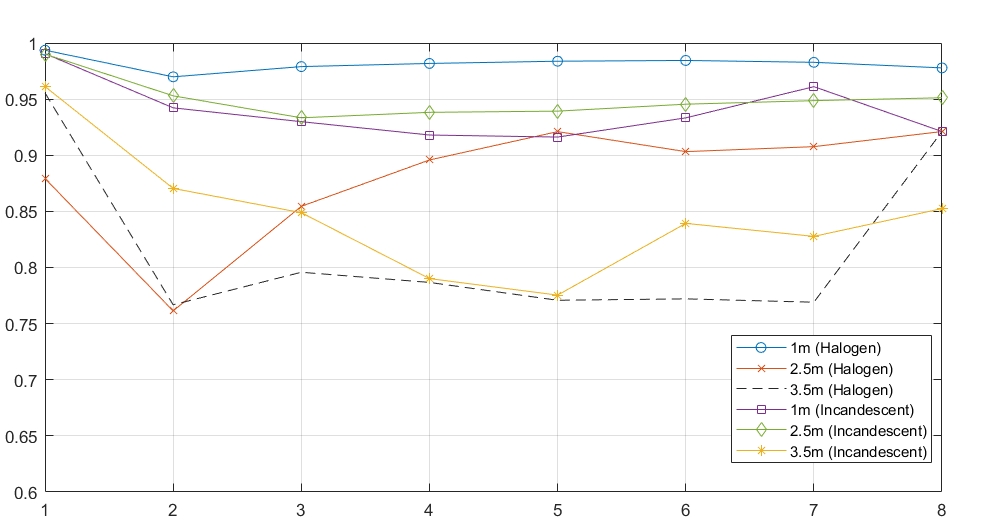}};
  \node[below=of img1, node distance=0cm, yshift=1cm,font=\color{black}] {Video duration (min)};
  \node[left=of img1, node distance=0cm, rotate=90, anchor=center,yshift=-0.7cm,font=\color{black}] {MCC};
\end{tikzpicture}
  \caption{\small MCC between the ENF estimated from a camera recording and the ENF measured at power mains for varying video duration and distance.}
  \label{video_distance_2lamps}
\end{figure}

Apart from the type of light source and the distance between the camera and the white wall, the duration of video recordings may vary. Since video duration is not known apriori, it is very important to challenge our ability to find a match in such circumstances. For that reason, we conducted another experiment that considered all three factors. A halogen lamp and an incandescent bulb were employed, while the camera was mounted at three different distances from the white wall background, i.e., 1 m, 2.5 m, and 3.5 m. Although the initial total duration of the recording was 8 min, we estimated the ENF for video durations $g=1,2, \ldots, 8$ min, as depicted in Figure \ref{video_distance_2lamps}. When the camera was mounted at a distance of 1 m, setups including either a halogen lamp or an incandescent bulb yielded high MCC values even for a short video duration of 1 min. Table~\ref{videodist} summarizes the MCC measurements. The top MCC for each light source is shown in boldface. Underlined MCC indicates the second and third best MCC for each pair of light sources and distance.

\begin{table}[!htb]
\centering
\caption{\sc {\small  MCC Between the ENF Estimated from Video and Power Mains for Varying Video Durations}}
\label{videodist}
\resizebox{\columnwidth}{!}{%
\begin{tabular}{l|llllllll}
\hline \hline
\textbf{Duration (min)} & \textbf{1} & \textbf{2} & \textbf{3} & \textbf{4} & \textbf{5} & \textbf{6} & \textbf{7} & \textbf{8} \\  \hline
1 m Halogen   & \textbf{0.9936} &  0.9699   &   0.9790  &  0.9818  &  0.9838 &  0.9845  &  0.9828  &  0.9778  \\
2.5 m Halogen & 0.8793          &  0.7613   &   0.8545  &  0.8958  &  \underline{0.9211} &  0.9032 &  0.9077 &
\underline{0.9211}\\
3.5 m Halogen   & \underline{0.9555} &     0.7668   &    0.7958   &  0.7867   &    0.7708    &  0.7720  &    0.7691    &   0.9211\\ \hline
1 m Incandescent     & \textbf{0.9905} &   0.9422   &   0.9299   &  0.9180 &  0.9162 &  0.9333 &  0.9611  &    0.9211\\
2.5 m Incandescent     & \underline{0.9902} &    0.9530   &   0.9334  &     0.9382   &    0.9393   &    0.9455   &  0.9487   &   0.9512\\
3.5 m Incandescent & \underline{0.9612} &   0.8704 &  0.8489 &  0.7901 &  0.7754 &  0.8393  &  0.8277 &    0.8525\\
 \hline \hline
\end{tabular}
}
\end{table}

To further assess the developed devices' performance and attest that the designed photodiode-based device delivers an accurate ground truth ENF, we used the same set of recordings against either the ENF extracted from power mains or the photodiode. Spectrum combining was used to estimate the ground truth ENF. The ENF estimation from video recordings was carried out by employing the BT spectral estimate. For varying distances between the camera and the white wall, as well as for both light sources, the same trend in MCC was observed in Table \ref{video_main_pd_BTtable}.

\begin{table}[!htb]
\centering
\caption{\sc {\small MCC Between the ENF Estimated from a Video Recording and Either the Power Mains or the Photodiode-device Output}}
\label{video_main_pd_BTtable}
\resizebox{\columnwidth}{!}{%
\begin{tabular}{l|lllllll}
\hline \hline
\textbf{Distance (m)} & \textbf{0.5} & \textbf{1} & \textbf{1.5}  & \textbf{2} & \textbf{2.5} & \textbf{3} & \textbf{3.5} \\  \hline
Mains (Halogen)             & 0.9888   & 0.9989    & 0.9904 &  0.9928  & 0.9968  & 0.9932 &    0.9344  \\
Photodiode (Halogen)        & 0.9      & 0.9998    & 0.999  &  0.9996  & 0.9998  & 0.9995 &    0.9445  \\ \hline
Mains (Incandescent)        & 0.9938   & 0.9983    & 0.8892 &  0.9949  & 0.9977  & 0.99   &    0.9944 \\
Photodiode (Incandescent)   & 0.9999   & 0.9998    & 0.8905 &  0.9998  & 0.9999  & 0.9997 &    0.9991  \\
 \hline \hline
\end{tabular}
}
\end{table}

\begin{figure}[!ht]
   \centering
  \begin{tikzpicture}
  \node (img1)  {\includegraphics[width=90mm,scale=0.85]{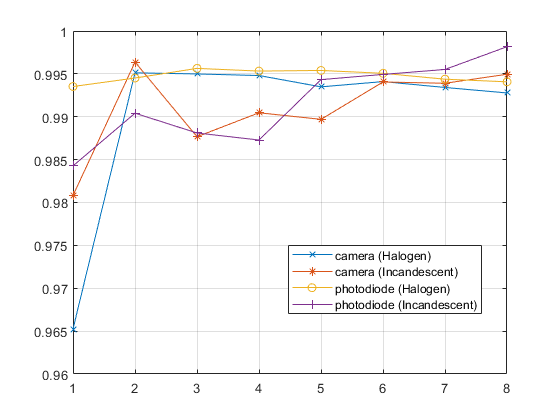}};
  \node[below=of img1, node distance=0cm, yshift=1cm,font=\color{black}] {Video duration  (min)};
  \node[left=of img1, node distance=0cm, rotate=90, anchor=center,yshift=-1cm,font=\color{black}] {MCC};
\end{tikzpicture}
  \caption{\small MCC between the ENF estimated from both optical sensors' recordings and power mains.}
  \label{video_vs_pd}
\end{figure}

The ENF estimated at the output of the photodiode-based device was compared to the ENF measured at power mains. Spectrum combining was used in both cases. The ENF was also extracted from the recording captured by the camera, which was placed  $2$ m far away from the white wall for both illumination sources and various video durations using SLIC and STFT.
The top MCC value of 0.9964 was observed between the ENF extracted from a 2-minute video when an incandescent bulb illuminated the scene and that measured from power mains. Regarding the photodiode-based device, the top MCC value of 0.9982 was observed between the ENF extracted from an 8-minute recording when an incandescent bulb illuminated the scene and that measured from power mains.
The MCC is plotted for various recording durations of the video and the photodiode signals in Figure~\ref{video_vs_pd}.

 The MCC measurements are listed in Table~\ref{video_vs_pd1}. Top MCCs are shown in boldface for each optical sensor.
\begin{table}[!htb]
\centering
\caption{\sc {\small Effect of Various Durations and Two different Light Sources in MCC}}
\label{video_vs_pd1}
\resizebox{\columnwidth}{!}{%
\begin{tabular}{l|llllllll}
\hline \hline
\textbf{Duration (min)} & \textbf{1} & \textbf{2} & \textbf{3} & \textbf{4} & \textbf{5} & \textbf{6} & \textbf{7} & \textbf{8} \\  \hline
\textbf{Camera (Halogen)}           & 0.9652   & 0.9951          &  0.9950 & 0.9948  & 0.9935 & 0.9941 &    0.9934 &  0.9928 \\
\textbf{Camera (Incandescent)}      & 0.9808   & \textbf{0.9964} & 0.9877   & 0.9905 & 0.9897  & 0.9941 &  0.9939  &  0.9950\\ \hline
\textbf{Photodiode (Halogen)}       & 0.9935    & 0.9945          & 0.9956  &  0.9953 &  0.9954 & 0.9951 & 0.9944 & 0.9941 \\
\textbf{Photodiode (Incandescent)}  & 0.9843   & 0.9904 & 0.9881  & 0.9873 &  0.9943 & 0.9949 & 0.9955  & \textbf{0.9982}\\
 \hline \hline
\end{tabular}
}
\end{table}

Furthermore, it is demonstrated that the ENF measured at the output of the photodiode-based device can be employed as a ground truth ENF.
The MCC measurements between the ENF estimated from the photodiode-based device and the power mains using spectrum combining are reported. Seven recordings of 8-minute duration each were captured. When a halogen lamp was employed, the top measured MCC was  0.9988, while the top MCC was 0.9989 for the incandescent bulb. For all recordings, the MCC  exceeded 0.99 regardless of the light source illuminating the scene, as seen in Table~\ref{spectrumMCC}.
\begin{table}[!htb]
\centering
\caption{\sc{\small MCC Between the ENF Extracted at the Output of the Photodiode device and the Power Mains}}
\label{spectrumMCC}
\resizebox{0.5\columnwidth}{!}{%
\begin{tabular}{l|ll}
\hline \hline
\textbf{\# Recording} & \textbf{Halogen} & \textbf{Incandescent} \\  \hline
1   &   0.9906          & 0.9946     \\
2   & \textbf{0.9988}   & \textbf{0.9989}      \\
3   &  0.9904           & 0.9975 \\
4   &  0.9941           & 0.9982 \\
5   &  0.996            & 0.9979\\
6   &  0.9934           & 0.9918 \\
7   &  0.9935           & 0.9935 \\
 \hline \hline
\end{tabular}
}
\end{table}

\subsection{Second set of experiments}

\begin{figure}[!ht]
   \centering
  \scalebox{1.0}{\includegraphics[width=80mm,scale=0.8]{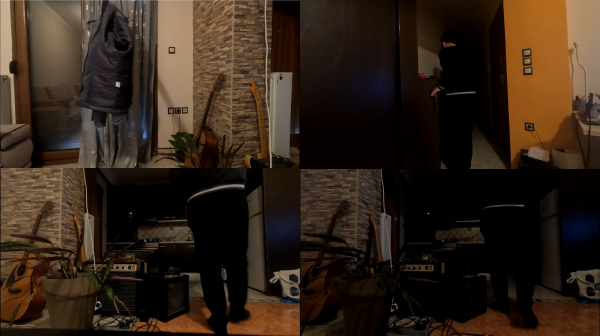}}
  \caption{\small Snapshots from the 4 non-static videos of the second set of experiments.}
  
  \label{thumbnail}
\end{figure}

A  challenging set of experiments was conducted to test the ability to timestamp non-static video recordings whose snapshots are shown in Figure~\ref{thumbnail}. Top left: Non-static video of a scene with various textures and reflection coefficients illuminated by a halogen lamp, where a moving jacket is introduced to the scene, referred to as video (a). Top right: A dimly lit hallway illuminated by a halogen bulb displaying human activity of varying obstruction, referred to as video (b). Bottom left: A video of a person moving in and out of a very complex scene illuminated by a halogen lamp in medium lighting condition, referred to as video (c). Bottom-right: The same scene as in the bottom-left when the lighting conditions are dimmer due to a low-power incandescent bulb, referred to as video (d). Each recording had a duration of 10 minutes.

The ENF estimation from video recordings was carried out by applying the SLIC-based approach~ \cite{Vatansever2017}~\cite{Karantaidis2021b}, bandpass filtering around the aliased light flickering captured by the camera for various filter orders as was set in each experiment, and ESPRIT unless otherwise stated.
Using spectrum combining, comparisons were made against the ENF estimated from either the power mains or the photodiode.
The objective is to find the best parameters yielding a satisfactory MCC compared with the ground truth from both the mains power and the photodiode.

For video (a), the experiment took place in a well-lit environment with many textures. A jacket moved in front of the camera throughout the video, not causing sudden changes in the overall light intensity. A high MCC of 0.997 was measured using SLIC, bandpass filtering with pass-band [9.9, 10.1] Hz of order $\nu$ = 111, and STFT for segment duration of 21s.

\begin{figure}[!ht]
   \centering
  \begin{tikzpicture}
  \node (img1)  {\includegraphics[width=90mm,scale=0.9]{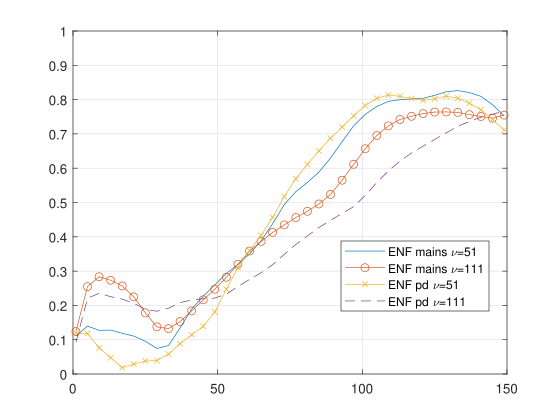}};
  \node[below=of img1, node distance=0cm, yshift=1cm,font=\color{black}] {Duration (min)};
  \node[left=of img1, node distance=0cm, rotate=90, anchor=center,yshift=-1.2cm,font=\color{black}] {MCC};
\end{tikzpicture}
  \caption{\small MCC between the ENF estimated from video (b) and either ground truth (mains or photodiode).} 
  \label{exp_b}
\end{figure}
Video (b) involves a person moving in and out of the frame in a dim environment. Despite the fact that more than half of the scene consists of dark objects with low reflection coefficients, an adequate MCC of 0.8133 and 0.8131 was obtained, respectively, when the ENF estimated from the video was compared against the ENF extracted from power mains or that captured by the photodiode-based sensing device. Figure~\ref{exp_b} depicts the MCC between the ENF estimated from the video and the ground truths captured by mains or photodiode for varying durations. The pass-band used in the experiment was [9.99, 10.19] Hz. It is shown that the best results are acquired for a segment duration of 133 s when a bandpass filter order of $\nu$=51 was employed. When the ground ENF was captured by the photodiode, the highest MCC value was observed for a segment duration of 109 s.

\begin{figure}[!ht]
   \centering
  \begin{tikzpicture}
  \node (img1)  {\includegraphics[width=90mm,scale=0.9]{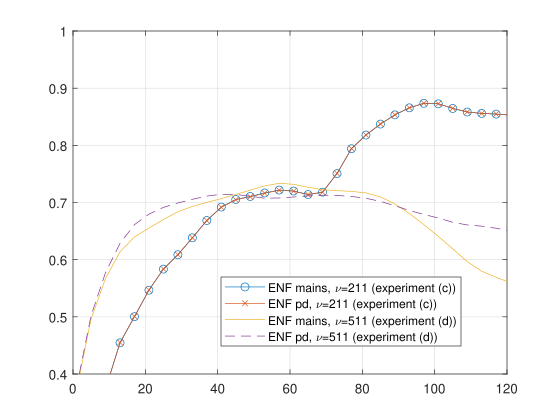}};
  \node[below=of img1, node distance=0cm, yshift=1cm,font=\color{black}] {Duration $g$ (min)};
  \node[left=of img1, node distance=0cm, rotate=90, anchor=center,yshift=-1.2cm,font=\color{black}] {MCC};
\end{tikzpicture}
  \caption{\small MCC of the ENF estimated from the videos (c) and (d) against the ground truths (mains and photodiode).}
  \label{exp_c_d}
\end{figure}
The final two experiments involving videos (c) and (d)  are the most challenging, containing a great number of objects, different surfaces, and shadows. In the former, the scene was illuminated by a halogen lamp. Applying SLIC and a bandpass filter of order $\nu$=211 before ESPRIT was used for ENF estimation, an MCC of 0.8736 was obtained for both ground truths. In the latter, we expected to acquire a lower MCC due to the lower overall light intensity fluctuation in proportion to the obstruction the moving person causes to the scene. The procedure is as above, with the difference that a bandpass filter of an order of $\nu$=511 was used. An MCC of 0.7336 was obtained when the ENF extracted from the video was compared to the ENF from power mains, while an MCC of 0.7143 was measured when compared to the ENF from the photodiode. Figure~\ref{exp_c_d} depicts the MCC obtained in both cases as a function of the segment duration. It is seen that the MCC between the ENF estimated from the video and either ground truth ENF follows the same trend for video (c). In contrast, the MCC is slightly higher when the comparison is made against the ENF from the power mains than the ENF from the photodiode. The pass-band in experiments (c) and (d) was [9.99, 10.19] Hz and [10.04, 10.14] Hz, respectively.

\section{Conclusions}
\label{sec:conc}

An optical sensor based on a photodiode has been developed to capture light intensity variations and enable ENF extraction in real-life scenarios. Multiple experiments have been conducted when a camera is used as a second optical sensor to record static and non-static videos.
Experiments have attested that the ENF estimated from the photodiode device using the spectrum combining approach can be employed as a reference signal for either a halogen or an incandescent bulb. The MCC between the ENF estimated from a camera recording and the ENF estimated from either the power mains or the photodiode-device output follows the same trend, confirming that the photodiode-device can provide a reliable reference ENF signal. 

It has been demonstrated that collecting a valid ground truth is possible without needing a device plugged into power mains. This fact allows battery-powered devices to be used as a means to extract ENF.
Future research would address ENF estimation in indoor environments where LED bulbs illuminate a scene. LED lighting is now more than ever commercially available, replacing older lighting technology such as incandescent bulbs at an increased rate. Therefore, it is urgent to challenge the ability to timestamp video recordings in such environments.

\section*{Acknowledgements}

G. Karantaidis' research work was supported by the Hellenic Foundation for
Research and Innovation (HFRI) under the HFRI PhD Fellowship grant  (Fellowship Number: 820). C.
Kotropoulos’ research was funded by the Hellenic Foundation for Research
and Innovation (HFRI) (Faculty Support Code: 388).

\bibliographystyle{IEEEtran}
\bibliography{strings}

\end{document}